%% file: main.tex
\documentclass{article}

\usepackage{arxiv}
\input{macros}
\usepackage[square,sort,comma,numbers]{natbib}
\usepackage[utf8]{inputenc} 
\usepackage[T1]{fontenc}    
\usepackage{hyperref}       
\usepackage{url}            
\usepackage{booktabs}       
\usepackage{amsmath,amssymb,amsfonts}       
\usepackage{nicefrac}       
\usepackage{microtype}      
\usepackage{graphicx}
\usepackage{algorithmic}
\usepackage{textcomp}
\usepackage{xcolor}
\usepackage[noend, linesnumbered, ruled]{algorithm2e}
\renewcommand{\algorithmcfname}{ALGORITHM}
\SetKwIF{If}{ElseIf}{Else}{if}{}{else if}{else}{end if}%
\SetKwFor{While}{while}{}{end while}%
\SetKwFor{For}{for}{}{end for}
\SetKwRepeat{Do}{do}{while}

\title{Learn to Earn: Enabling Coordination Within a Ride-Hailing Fleet}

\author{Harshal A. Chaudhari \\
	Department of Computer Science\\
	Boston University\\
	Boston, USA \\
	\texttt{harshal@cs.bu.edu} \\
	\And
	John W. Byers \\
	Department of Computer Science\\
	Boston University\\
	Boston, USA\\
	\texttt{byers@cs.bu.edu} \\
	 \And
	Evimaria Terzi \\
    Department of Computer Science\\
	Boston University \\
	Boston, USA\\
	\texttt{evimaria@cs.bu.edu} \\
}

\hypersetup{
pdftitle={Learn to Earn: Enabling Coordination Within a Ride-Hailing Fleet},
pdfsubject={q-bio.NC, q-bio.QM},
pdfauthor={Harshal A. Chaudhari, John W. Byers, Evimaria Terzi},
pdfkeywords={ride-hailing, reinforcement learning, combinatorial methods},
}

\begin{document}
\maketitle

\input{abstract}
\input{introduction}
\input{related_work}

\input{problem_setup}
\input{approach}
\input{experiments}
\input{conclusion}
\input{acknowledgement}

\bibliographystyle{unsrt}
\bibliography{bibliography}
\end{document}

%% file: macros.tex
\usepackage{amsmath,amsfonts,bm}

\newcommand{\hexbins}{{{\ensuremath{\mathbb{H}}}}}
\newcommand{\timesteps}{{{\ensuremath{\mathbb{T}}}}}

\newcommand{\relocatingdrivers}{{{\ensuremath{\mathbb{R}}}}}
\newcommand{\waitingdrivers}{{{\ensuremath{\mathbb{W}}}}}

\newcommand{\demandmatrix}{{{\ensuremath{\mathbf{D}}}}}
\newcommand{\traveltimematrix}{{{\ensuremath{\mathbf{T}}}}}
\newcommand{\rewardmatrix}{{{\ensuremath{\mathbf{R}}}}}
\newcommand{\actionmatrix}{{{\ensuremath{\mathbf{A}}}}}
\newcommand{\imbalancematrix}{{{\ensuremath{\Delta}}}}
\newcommand{\rebalancematrix}{{{\ensuremath{\mathbf{Z}}}}}

\newcommand{\policy}{{{\ensuremath{\pi}}}}
\newcommand{\coordination}{{{\ensuremath{\xi}}}}
\newcommand{\rebalancinggraph}{{{\ensuremath{\mathcal{G}}}}}
\newcommand{\imbalancethreshold}{{{\ensuremath{\Lambda}}}}

\DeclareMathOperator*{\argmax}{arg\,max}

\newcommand{\spara}[1]{\smallskip\noindent{\bf{#1}}}
\newcommand{\mpara}[1]{\medskip\noindent{\bf{#1}}}

\newcommand{\squishlist}{\begin{list}{$\bullet$}
    {
    \setlength{\itemsep}{0pt}
    \setlength{\parsep}{3pt}
    \setlength{\topsep}{3pt}
    \setlength{\partopsep}{0pt}
    \setlength{\leftmargin}{1.5em}
    \setlength{\labelwidth}{1em}
    \setlength{\labelsep}{0.5em}
    }
}

\newcommand{\squishend}{\end{list}}

%% file: abstract.tex

\begin{abstract}
The problem of optimizing social welfare objectives on multi-sided ride-hailing 
    platforms such as Uber, Lyft, etc., is challenging, due to misalignment
    of objectives between drivers, passengers, and the platform itself.
An ideal solution aims to minimize the response time for each hyperlocal 
    passenger ride request, while simultaneously maintaining high demand
    satisfaction and supply utilization across the entire city. 
Economists tend to rely on dynamic pricing mechanisms that stifle
    price-sensitive excess demand and resolve the supply-demand imbalances 
    emerging in specific neighborhoods.
In contrast, computer scientists primarily view it as a demand prediction 
    problem with the goal of pre-emptively repositioning supply to such
    neighborhoods using \emph{black-box} coordinated multi-agent 
    deep reinforcement learning-based approaches.
Here, we introduce explainability in the existing supply-repositioning 
    approaches by establishing the need for coordination between the drivers 
    at specific locations and times.
Explicit need-based coordination allows our framework to use a
    simpler \emph{non-deep} reinforcement learning-based approach,
    thereby enabling it to explain its recommendations \emph{ex-post}.
Moreover, it provides \emph{envy-free} recommendations i.e., drivers
    at the same location and time do not envy one another's future earnings.
Our experimental evaluation demonstrates the effectiveness, the robustness, and 
    the generalizability of our framework. 
Finally, in contrast to previous works, we make available a reinforcement 
    learning environment for \emph{end-to-end} reproducibility of our work and 
    to encourage future comparative studies.
\end{abstract}

%% file: introduction.tex
\section{Introduction}
\label{sec:introduction}

Popular ride-hailing platforms such as Uber, Lyft, Didi Chuxing, Ola, etc., have
    revolutionized the daily commute in cities across the world.
Globally valued at over \$61 billion and expected to grow \$218 billion by
    2025
    these platforms operate as multi-sided marketplaces, seamlessly connecting drivers
    with riders through their smartphone applications~\cite{noauthor_undated-gj}.
The explosive growth of these ride-hailing platforms has motivated a wide array of 
    questions for academic research at the intersection of computer science 
    and economics, as we discuss in the related work section.

A large portion of this work aims to improve the performance of the platforms 
    ensuring high-reliability service for the passengers and high utilization 
    and earnings for the drivers. 
The two main thrusts are \emph{dynamic pricing} and \emph{capacity repositioning}. 
Dynamic pricing
    ~\cite{Yan2018-wq, Besbes2019-ds, Banerjee2015-jx, Castillo2018-he, Garg2019-vs} 
    aims to balance demand and supply by increasing prices in certain
    neighborhoods. 
Intuitively, the increased prices curtail price-sensitive demand and assist the 
    platform in ensuring a high-reliability service.
On the flip side, the potential for higher earnings also encourages more drivers 
    to join the platform during such ``price surges". 
The dynamic pricing literature uses  game-theoretic analyses of the 
    ride-hailing markets to shove its 
    effectiveness as a platform control mechanism. 

Capacity repositioning approach aims to improve the platforms' performance
    by assisting drivers with recommendations for relocations inside a city. 
Although the initial work in this domain has focused on modeling the 
    driver-repositioning problems as combinatorial optimization
    problems~\cite{Lee2004-hh, Zhang2016-vz, Seow2010-qg,Xu2018-xb, Zhang2017-id}, 
    the need to deal with large driver fleets and the availability of 
    high-dimensional historical data has recently led to the development of 
    machine learning methods for the same problem~\cite{Mnih2013-sj,Tang2019-xu,
    Lin2018-vs, Wen2017-vp, Wang2018-bv}. 
Such approaches predominantly use multi-agent coordination reinforcement
    learning to solve a global capacity repositioning problem, using various 
    forms of \emph{attention mechanisms} in neural netowrks to achieve coordination.
By design, they make a key 
    assumption that there is \emph{always} a need for coordination in the market. 
This assumption necessitates them to leverage recent breakthroughs
    in the scalability of deep learning models to exploit the high-dimensional
    historical data.
Deep-learning based methods have a
    large number of hyperparameters required in their training, making their 
    performance susceptible to external perturbations.
Moreover, the \emph{black-box} coordination issue, as yet unresolved, is a
    potential liability when using deep-learning based systems in domains such
    as this, where platform controllers would like to understand how the
    choices of recommendations to human drivers were made.

We aim to revise the capacity-repositioning approach by relaxing the assumption 
    that coordination is always necessary.  
Specifically, our approach leverages the
    observation that driver actions are independent
    at most times of the day, with coordination required only during periodic
    times of peak demand, such as a rush hour. 
Furthermore, the instances of supply-demand imbalances in a city are usually 
    restricted to distinct neighborhoods.
We exploit this loose spatio-temporal coupling of supply and demand to
    learn \emph{when} and \emph{where} the drivers need to coordinate, and 
    otherwise act independently for the rest of the time.
This observation allows us to combine vanilla reinforcement learning 
    (i.e., not deep learning) algorithms with simple combinatorial techniques 
    for solving the repositioning problem.
Moreover, our framework is scalable because the 
    sizes of the combinatorial problems we need to solve in order to 
    achieve capacity repositioning are constrained by the number of 
    imbalanced neighborhoods.
Broadly, our model is a combination of the 
    combinatorial and machine-learning approaches to capacity repositioning.

As our framework does not rely on deep learning, we are able to explain 
    \emph{ex-post} all the recommendations given to
    the drivers, taking a step in direction of \emph{transparent} A.I. proposed
    in the recent \emph{General Data Protection Regulations} (GDPR)
    guidelines~\cite{Vincent2019-eg}.
Moreover, our approach is \emph{envy-free} in the sense that drivers at the same 
    location and time do not envy one another's future earnings.
The resulting model is relatively parameter-free and hence generalizes well in
    presence of daily variations in supply and demand.
Finally, our framework is amenable to 
    integration with any dynamic-pricing models by easily augmenting our data
    with the effects of such a model.

\spara{Contributions:} To summarize, the contributions of our paper are the following:
\squishlist
\item We consider the problem of capacity relocation on a 
        ride-hailing platform in order to maximize welfare and propose a robust, 
        explainable and scalable framework that combines simple combinatorial techniques
        with vanilla reinforcement learning algorithms.
\item We perform a thorough experimental evaluation of the dynamics of
        the fleet-management system and the effectiveness, robustness to imperfect
        hyperparameter tunings, and 
        generalizability of our models in the presence of external perturbations.
\item We also make available an OpenAI gym environment\footnote{OpenAI
        Gym is a toolkit for developing and comparing reinforcement learning
        algorithms~\cite{open-ai-gym}.}
        named \texttt{nyc-yellow-taxi-v0} at \cite{github-page} so that any 
        future multiagent reinforcement learning 
        algorithm can be easily applied to this problem.
    To the best of our knowledge, this is the first environment based on 
        real-world datasets.
\squishend

%% file: related_work.tex
\section{Related Work}
\label{sec:related_work}

In this section, we discuss existing work in comparison to ours.

\spara{Driver recommender systems:}
The problem of spatiotemporal demand prediction 
    to inform an idle taxi driver of favorable locations for passenger
    pickups had been studied extensively even before the advent of ride-hailing
    services. 
These works focus on the case of a self-interested individual acting in isolation.
For instance, Li et al.~\cite{Li2011-qm} use
    large-scale taxi GPS trace dataset to identify salient features associated
    with successful passenger pickup locations, while in two separate studies, 
    Yuan et al.~\cite{Yuan2011-pd, Yuan2013-gj} develop a recommender
    system to guide both idle taxi drivers and waiting passengers to convenient
    locations in order to optimize social welfare.
More recently, Chaudhari et al.~\cite{Chaudhari2018-cv} devise driver-oriented strategies to
    recommend favorable driving schedules and pickup locations
    to optimize the earnings of an individual on-demand ride-sharing driver.
In contrast to our approach, the recommender systems in these works 
    are agnostic to driver interactions
    and may result in unfavorable supply excesses in certain locations
    when adopted by many drivers simultaneously.

\spara{Capacity repositioning systems:}
Traditional works in driver dispatch systems~\cite{Lee2004-hh, Zhang2016-vz,
    Seow2010-qg} typically rely on queuing-theoretical models to asymptotically
    optimize supply-demand matching while reducing congestion-related issues.
Aided by real-time data processing leading to precise estimates of supply and
    demand, recent approaches~\cite{Xu2018-xb, Zhang2017-id} 
    draw upon combinatorial optimization, leveraging
    demand volume and ride destination forecasting. 
These approaches do not scale well to use cases on contemporary platforms,
    where fleets of as many as 10,000 drivers serve a single city.

More recent work~\cite{Mnih2013-sj,Tang2019-xu,
    Lin2018-vs, Wen2017-vp, Wang2018-bv} 
    addresses the scalability issue by using deep reinforcement learning to learn 
    control policies in high-dimensional input space. 
While effective in this high-volume data domain, these methods rely on
    external proprietary models to generate inputs for the driver dispatch
    systems.
For example, the approach of Kaixiang et al.~\cite{Lin2018-vs} heavily relies on 
    a proprietary simulator built by Didi Chuxing in order to generate inputs 
    required during model training, making it impossible 
    to reproduce their results for comparison purposes.
To the best of our knowledge, building such a simulator is itself an 
   active research problem.
Moreover, deep-learning based techniques suffer from a lack of
    explainability.
Cognizant of these issues, our approach does not rely on any proprietary models 
    but rather learns high-quality solutions \emph{from scratch} based solely 
    upon historically observed data.
Moreover, we achieve that without sacrificing the explainability of the
    model.
In the absence of the need of coordination, our model assumes homogeneity of 
    drivers in same location and provides
    \emph{envy-free} recommendations while also making it scalable. 
Furthermore, we have made publicly available our entire codebase and
    reinforcement learning environment required to reproduce every result 
    presented in this work while also
    enabling comparative future studies~\cite{github-page}.

From the learning point of view, our approach not only learns \emph{how} to coordinate 
    but also \emph{when} it is required to do so. 
This is achieved by augmenting  vanilla reinforcement 
    learning (in the form of tabular Q-learning) with
    combinatorial techniques to aid the rebalancing of driver
    distribution.

\spara{Platform economics:}
Studies of ride-hailing services as multi-sided economic marketplaces have 
    investigated the impacts of the platform's pricing policies on the platform profits, 
    the consumer surplus, and the driver
    wages~\cite{Castillo2018-he, Bimpikis2016-yf, Besbes2019-ds, Ma2018-hb}.
S{\"u}hr et al.~\cite{Tom_Suhr-ps} investigate fairness in driver earnings 
    distribution using driver-passenger matchings optimized to attain 
    income equality goals.
Recently, Chen et al.~\cite{Chen2019-li} combines platform economics with capacity
    repositioning problem using a contextual bandit framework. 
There is a growing body of literature studying the interplay between platform
    pricing and strategic driver behaviors, for which we refer the readers to
    ~\cite{Yan2018-wq}.
Our work contributes to this domain by 
    developing a scalable framework that can be used to verify the results of
    asymptotic dynamic pricing models via realistic simulations.

%% file: problem_setup.tex
\section{Problem Setup}
\label{sec:problem_setup}
In this section, we describe the basics of our problem setup and provide 
the necessary notation.

\subsection{City attributes}

Throughout the paper, we assume that the city is divided into a set of 
    $m$ non-overlapping hexagonal zones denoted by ${\hexbins}$. 
We also assume that time $t$ advances in discrete time steps i.e., 
    $\timesteps = \{1, \cdots, T\}$, a standard industry practice 
    described in \cite{Chen2019-ko}.
 
Finally, we assume a total of $n$ homogeneous drivers
    traveling between hexagon zones picking up and dropping off the passengers.

Our model uses the following city matrices and vectors that are time-varying; 
    i.e.,  their entries change at every time step. However, for notational 
    convenience, we do not introduce the time step $t$ subscript with the 
    entries unless required for context. 

\mpara{Demand matrix ({\demandmatrix}):}
A matrix $\demandmatrix \in \mathbb{R}^{m \times m}$ such that each entry
    $d(h, h')$ denotes the number of passengers requesting a ride from zone $h$ to
    zone $h'$ at time $t$. 
With appropriately sized hexagonal city zones, we find that
    $\forall h \in {\hexbins}, d(h, h)=0$.

\mpara{Travel time matrix ({\traveltimematrix}):}
A matrix $\traveltimematrix \in \mathbb{R}^{m \times m}$ such that each entry
    $\tau(h, h')$ denotes the number of discrete time steps required for transiting 
    from zone $h$ to zone $h'$. 

\mpara{Reward matrix ({\rewardmatrix}):}
A matrix $\rewardmatrix \in \mathbb{R}^{m \times m}$ such that each entry 
    $r(h,h')$ denotes the net reward for a taxi driver carrying a passenger 
    from zone $h$ to zone $h'$. 
The net rewards include driver's earnings for delivering the
    passenger at the destination minus the sundries such as gas cost, vehicle
    depreciation, etc. 
Hence, each entry of the matrix is of form $r(h, h') = \textrm{earnings}(h, h')
- \textrm{cost}(h, h')$.

\mpara{Driver actions ({\actionmatrix}):}
At each time step $t$, a driver in zone $h$ who is not currently on a
trip can choose one of the two actions.
\squishlist
    \item \textbf{Wait:} A wait action $a(h,h)$ involves waiting for a
            passenger in the current zone $i$ for the current time step. 
            If successful, it can lead to a trip to some other zone $h'$
            with the driver earning a reward of $r(h, h')$.  
        When the number of
            drivers choosing to wait in a zone exceeds the demand of the zone at
            the particular time, an unsuccessful wait may occur, and the driver earns
            a net reward of zero while staying in the same zone $h$ for the next time
            step.
    \item \textbf{Relocate:} A relocate action $a(h,h')$ involves
            relocation {\em without a passenger} from zone $h$ to zone $h'$. 
        Undertaking a relocate action costs a driver a value denoted 
            by $\textrm{cost}(h, h')$.
\squishend
Thus, we consider a total of $|\actionmatrix| = m^2$ actions.  
In case of a relocate action or a successful passenger pickup to zone $h'$, 
    the driver is busy traveling for next $\tau(h, h')$ time steps and is 
    presented with the next action choice at time $t + \tau(h, h')$, 
    while in case of an unsuccessful wait, the driver chooses the next action 
    at time $t + 1$.

\subsection{Model attributes}

Using the city attributes from the previous section, we now define the
    attributes of our model:

\mpara{Policy ({\policy}):}
A policy function $\policy : \hexbins \times \timesteps \rightarrow \actionmatrix$ 
    recommends the best action to drivers in every zone of the city at each time step,
    to maximize the model's objective function.  
We impose a constraint that all drivers in the same zone at the same time be 
    recommended the same action unless driver coordination is required to 
    resolve a supply-demand imbalance in the zone.  

A driver $i$ following a policy $\policy$ performs location 
    and time-dependent actions represented by a 3-tuple 
    $\phi^{\policy}_t(i) = (t, h, a)$, where $h$ and $a$ are the location of the driver
    and the action chosen at time $t$ respectively.
We assume that if a driver is busy at time $t$, the corresponding 3-tuple 
    is $(t, \varnothing,\varnothing)$.

\spara{Driver earnings ($\mathcal{E}$):}
Let function $E(t, h, a)$ denote the net earnings of a driver 
    on taking action $a$ at time $t$ while located in zone $h$. 
If the action leads a driver to zone $h'$, $E(t, h, a) = r(h, h')$. 
In the case of relocate action, the net earnings simply constitute the cost of 
    relocation i.e., 
    $r(h,h') = - \textrm{cost}(h, h')$.
We can denote the gross earnings of $n$ drivers following a policy $\pi$ by:
    \begin{equation*}
    \mathcal{E}^{\policy}(n, \demandmatrix, \traveltimematrix, \rewardmatrix) =
    \sum_{t=1}^{T} \sum_{i=1}^{n}  E\big(\phi^
    {\policy}_t(i)\big),
    \end{equation*}
    where $E(t, \varnothing, \varnothing) = 0$. 

\spara{Supply ($S$):}
A policy $\pi$ induces the movement of drivers between different city zones
    through action choices. 
The supply, i.e., the number of drivers at zone $h$ at time
    $t$ induced by a policy $\policy$ is denoted using the supply function
    $S^{\policy}(t, h)$.  

\spara{Demand fulfillment ($\mathcal{F}$):}
A driver in zone $h$ choosing the wait action $a(h,h)$ at time $t$ is randomly
    matched with any of the $\sum_{h'}d_t(h,h')$ passengers requesting a ride in zone
    $h$ at the same time.
Hence, a policy $\pi$, via its supply function, induces a demand fulfillment function. 
Demand fulfilled in zone $h$ at time $t$ when drivers follow a policy
    $\policy$ is denoted using the demand satisfaction function
    $F^{\policy}(t, h)$. 
Obviously, $\forall_{\policy \in \Pi}
    F^{\policy}(t, h) \leq \sum_{h'} d_t(h, h')$. 
Hence, total demand fulfilled over the course of time steps $t \in {\timesteps}$ 
    by $n$ drivers following the policy $\policy$ can be given by:
    \begin{equation*}
    \mathcal{F}^{\policy}(n,\demandmatrix, \traveltimematrix, \rewardmatrix) =
    \sum_{t=1}^{T} \sum_{h=1}^{m}  F ^{\policy}(t, h).
    \end{equation*}

\subsection{Problem statement}
Based on the above definitions, we now formulate the problem that we 
    solve. \\

\noindent\textsc{Problem 1(MaxEarnings):} Given time-varying matrices $\demandmatrix,
    \traveltimematrix, \rewardmatrix$ and the number of homogeneous drivers $n$,
    devise a policy $\policy^*$ such that
    \begin{equation}
    \label{eq:max_earnings_problem}
    \policy^* = \argmax_{\policy \in \Pi} \mathcal{E}^{\policy}(n,\demandmatrix,
    \traveltimematrix, \rewardmatrix).
    \end{equation}
Replacing the driver earnings ($\mathcal{E}$) by demand fulfillment
    ($\mathcal{F}$) in the Equation~\eqref{eq:max_earnings_problem} above
    results in a variant of \textsc{MaxEarnings} problem, in which the goal is to
    maximize fulfilled rides, referred to henceforth as
    the \textsc{MaxFulfillment} problem.

\subsection{Discussion}
\label{sec:problem_setup_discussion}
At a high level, our revenue maximization problem statement is similar to 
   existing work in the area. 
However, our modeling assumptions are significantly more realistic 
   than previous work, making our methods more amenable to 
   operational deployment.
For example, whereas our work assigns drivers to rides explicitly
   (and allocates rewards accordingly), previous work
   performs reward allocations proportionally, in a fluid model.
In our model, when two drivers compete for a single ride from $i$ to $j$, 
   one of the two driver gets the ride, gets paid, and ends up in $j$
   after a travel time $t$.
In the representative fluid model of~\cite{Lin2018-vs}, 
   both drivers get half of the payment, and two halves of
   drivers (conceptually) transit to $j$ in a fixed time step, 
   regardless of the distance traveled.
Although a fluid model such as this is tractable to solve for and optimize
   around, it has strong implications on the solution space, as it 
   notably removes time-dependent and driver-dependent features from the model.
Issues such as studying variance of driver earnings are not possible in these
   models, as all drivers starting at the same time and place will end up with 
   identical (quantized) trajectories and earnings.

We also note that in our framework, a wait action for an individual 
    driver is only successful if the driver is present in the same zone as
    the ride request. Hence, our framework cannot result into the 
    \textit{Wild Goose Chase (WGC)} phenomenon described by
    ~\cite{Castillo2018-he},
    in which high demand causes depletion of idle drivers on the streets,
    leading to suboptimal FCFS matches where drivers spend a significantly higher
    duration of time en route to pick up passengers.

%% file: approach.tex
\section{Learning Framework}
\label{sec:approach}
In this section, we describe our approach for solving the \textsc{MaxEarnings} 
    problem.
Our method is a \emph{model based reinforcement learning} approach, and its 
    description is provided in Algorithm~\ref{alg:general_learning}. 
 
As with any reinforcement learning based approach, we train our model by
    allowing the drivers to repeatedly interact with an environment 
    in form of the city's ride demand data from a representative day.
Each interaction, which is $T$ timesteps long, constitutes an \emph{episode}
    of the training process.
Each episode constitutes of 3 phases described below.

\input{algorithm}

\spara{Exploratory phase (lines 5-7): }
During this phase of the algorithm, drivers exhibit an
    \emph{exploratory behavior} by choosing a pseudo-random action 
    with a probability $\epsilon$.
These randomly chosen actions allow the model to explore a larger portion of
    the policy space, preventing its policy from converging to a local
    minimum.
This is similar to the $\epsilon$-greedy behavior of
    Q-learning~\cite{Sutton1998-wd}.

\spara{Exploitative phase (lines 9-12): }
During this phase of the algorithm, the rest of the drivers
    exhibit an \emph{exploitative behavior} using the policy learned up until 
    the previous episode of training.
The policy recommends exploitative actions to individual drivers based upon 
    the time of the day and their locations, independently of each other,
    henceforth referred to as \emph{independent actions}.
However, certain recommended actions may result in supply-demand imbalances
    when a large number of drivers relocate to the same city zone with an
    insufficient demand, or too few of them relocate to a zone with excess 
    demand.
We postulate that explicit coordination is essential to prevent such
    supply-demand imbalances from occurring.
Hence, we introduce the \emph{degree of coordination ({\coordination})} - a 
    probabilistic value that signifies the extent to which drivers located in 
    the same city zone need to coordinate their actions.
Whenever a zone has a positive degree of coordination, the exploitative actions
    recommended to a {\coordination} fraction of drivers in the zone are derived
    from solving a reward-maximizing linear program, henceforth referred to as
    \emph{coordinated actions}.

It should be noted that it is the explicit criterion for recommending a 
    coordinated action that sets our
    approach apart from recent works in the field of deep reinforcement
    learning across different applications and domains.

\spara{Learning phase (lines 15-18): }
Actions recommended in the exploratory and exploitative phases of
    the episode result in drivers picking up passengers or relocating themselves
    to different city zones, thereby observing rewards of their actions (line 13).
The learning phase of the algorithm computes a rebalancing matrix (line 14) to use
    in conjunction with the observed rewards to further improve upon the
    policy.

Having developed an intuition for the major building blocks of
    Algorithm~\ref{alg:general_learning}, we now explain these phases in 
    greater detail. 

\subsection{Exploratory phase}
Over the course of training, when a driver located in 
    zone $h$ chooses to explore, we model the probability of driver's 
    exploratory ride distance using a Gaussian function with a random variable 
    $K_{\geq 0}$. 
Specifically, the probability that a driver relocates to a zone at distance 
    $k \geq 0$ is given by:
    $Pr[K = k] = be^{-\frac{k^2}{2c^2}}$.

After sampling an exploration distance $k$, the driver chooses the actual 
    destination by sampling uniformly at random from all the zones at the 
    distance $k$. 
When $k=0$, the driver chooses to wait in the current zone, while for $k >0$, 
    the driver chooses a relocate action. 
The experiments in this paper were all conducted using $b=0.7$ and
    $c=1$ (chosen via grid-search), allowing explorations upto 3 hexagonal zones away.
In contrast,~\cite{Lin2018-vs} restricts drivers to single zone distance
    relocations, reducing their ability to learn policies that mitigate 
    supply-demand imbalances by relocating supply from zones further away in a
    single timestep.
Over the course of training, $\epsilon$ is annealed exponentially from
    1 to 0, thereby outputting an entirely exploitative model at the end of
    the training.

\subsection{Exploitative phase}
Exploitative behavior is manifested in the form of independent
    actions (line 10) and coordinated actions (line 12) when the degree of
    coordination is positive.
We detail these next.

\subsubsection{Choice of independent action}
For each independent action chosen by a driver we record the associated 
    reward earned. 
The reward earned is then used to update the \emph{value} of the action for the 
    next episode, based on the learning rate ($\alpha$) and the
    discount factor ($\gamma$). 
These values are stored in a Q-table denoted by 
    $Q_I \in \mathbb{R}^{T \times m \times |\textbf{A}|}$.
For each zone $h$, at time $t$, the best independent action ($a^*$) in line 10 of 
    Algorithm~\ref{alg:general_learning} is chosen by
    \begin{equation*}
        a^*(t, h) = \argmax_{a \in \actionmatrix_h}Q_I(t, h, a), 
    \end{equation*}
    where $\actionmatrix_h$ refers to the $h$-th row of {\actionmatrix}.

\subsubsection{Independent learning}
Based upon the observations of drivers undertaking independent actions (both
    exploratory and exploitative), we
    update the independent learning matrix ($Q_I$) as
    described below.


\spara{Updating $Q_I$ for wait actions:}
Let $\waitingdrivers_{(h, h')}$ denote the number of drivers choosing to wait
    in zone $h$ at time $t$, and ending up in zone $h'$. 
A successful wait generates net earnings $E\big(t, h, a(h, h)\big) = r(h, h')$ 
    and consumes a travel time $\tau(h, h')$, while 
    an unsuccessful wait i.e., $h' = h$, generates zero net earnings
    and consumes one timestep.
The utility of the wait action is therefore
    \begin{equation*}
        \mathcal{U}_{(t, h,h)} = \sum_{h'} 
        \waitingdrivers_{(h, h')}
        \bigg[E\big(t, h,
            a(h,h)\big) + \gamma 
    Q_I(t', h', a^*(t', h'))\bigg],
    \end{equation*}
    where $t' = t + \tau(h, h')$ and we discount the future rewards with 
    a factor $\gamma$.
We use the utility of the wait action to update the entry $Q_I(t, h, h)$ as follows: 
    \begin{equation}
    \label{eq:Q_I_update_wait}
        Q_I(t, h, h) \leftarrow (1 - \alpha)Q_I(t, h, h) +
        \frac{\alpha}{\sum_{h'}\waitingdrivers_{(h, h')}}\mathcal{U}_{(t, h,h)}.
    \end{equation}
Normalizing the update term by the number of drivers choosing the wait action 
    captures the average utility of the wait action.
The term $Q_I(t, h, h)$ on the right hand side of the equation denotes the 
    values learned upto the previous episode of the training, and $\alpha$ is 
    the learning rate.

\spara{Updating $Q_I$ for relocate actions:}
Let $\relocatingdrivers_{(h, h')}$ denote the number of drivers relocating from zone $h$
    to zone $h'$. The utility of such relocation is given by
    \begin{equation*}
        \mathcal{U}_{(t,h,h')} = \relocatingdrivers_{(h, h')} \bigg[E(t, h,
            a(h, h')) + \gamma
        Q_I(t', h', a^*(t', h'))\bigg],
    \end{equation*}
    where $t' = t + \tau(h, h')$. 
We use the utility of the relocate actions to update the entry $Q_I(t, h, h')$
    of the independent table as follows: 
    \begin{equation}
    \label{eq:Q_I_update_relocate}
        Q_I(t, h, h') \leftarrow (1 - \alpha)Q_I(t, h, h') 
        + \frac{\alpha}{\relocatingdrivers_{(h, h')}}\mathcal{U}_{(t, h,h')}.
    \end{equation}
Using Equations~\eqref{eq:Q_I_update_wait} and ~\eqref{eq:Q_I_update_relocate}, the
    $Q_I$ matrix is updated in line 16 of Algorithm~\ref{alg:general_learning} 
    using the evidence obtained via simulations in form of 
    utilities $\mathcal{U}_{(t, h, h')}$ of both the wait and relocate actions.

\subsubsection{Choice of coordinated action}
The choice of coordinated action is more intricate and non-standard, and we 
    next explain it in detail.
To guide the coordinated behavior of drivers in line 12 of
    Algorithm~\ref{alg:general_learning}, we solve a reward-maximizing
    rebalancing operation between city zones experiencing supply-demand
    imbalances. 
There are two principal components driving the coordinated behavior: 
\emph{degree of coordination} ({\coordination}) which controls the need of 
    coordination in a particular zone at a time, and \emph{coordinated learning
    matrix} ($Q_C$) which determines the choice of action as a response to the 
    need of coordination.
Thus, each coordinated action is associated with a probability for it to participate in
    the rebalancing operation that is stored in the matrix $Q_C$. 
Note that $Q_C$ contains learned probabilities, as against the usual 
    action-value nature of $Q_I$. 

Let the policy learned at the end of $k$-th episode during training be denoted
    by ${\policy}_k$. 
Following this policy induces a driver supply $S^{\policy_k}$ during 
    the $(k+1)$-th episode of training. 
For each zone $h$, at time $t$, the coordinated action ($a^c$) in line 12 of 
    Algorithm~\ref{alg:general_learning} is obtained by uniformly sampling 
    from the probability vector $Q_C(t, h)$.

\spara{Imbalance matrix (\imbalancematrix):}
A matrix ${\imbalancematrix} \in \relocatingdrivers^{|\timesteps| \times
    m}$ such that each entry $\delta(t,h)$ denotes the supply-demand imbalance
    experienced at zone $h$ at time $t$ during the $(k+1)$-th episode. 
Specifically, each entry of the imbalance matrix can be given by, $\delta(t, h) =
    S^{\policy_k}(t, h) - \sum_{h'} d_t(h, h')$.
We mask the imbalance matrix using an imbalance threshold parameter 
    {\imbalancethreshold} such that,
    \begin{equation*}
        \delta(t,h) = 
            \begin{cases}
            \delta(t, h) &\text{ if } \big|\delta(t, h)\big|
            \geq \imbalancethreshold \\
            0 &\text{ otherwise. }
        \end{cases}
    \end{equation*}
Using this parameter allows us to control the level of imbalances that the
    framework should attempt to mitigate. 

\spara{Rebalancing graph (\rebalancinggraph):}
Based upon the supply-demand imbalance matrix induced at the end of an
    episode, we create the \emph{rebalancing graph}
    $\rebalancinggraph = (V,E)$ consisting of
    imbalanced zones as nodes and edges as corresponding relocation actions between
    them. 
This is a bipartitle graph with nodes set $V = \{V_+ \cup V_-\}$ where $V_+$ is the set
    of nodes with excess supply i.e., $\delta(t,h) > 0$ and $V_-$ is the
    set of nodes with supply deficit i.e., $\delta(t, h) < 0$.
Thus each node $v_i \in V$ in the rebalancing graph is associated with 
    three attribues: \textit{imbalanced zone} $(v_i^h)$, \textit{time of
    imbalance} $(v_i^t)$ and \textit{magnitude of imbalance} $\big(\delta(v_i^t, v_i^h)\big)$.
The edge set $E$ consists of directed edges from the nodes in $V_+$ 
    to nodes in $V_-$ and they model feasible relocations. 
Thus:
    $E = \bigg\{ e_{ij}: v_i \in V_+, v_j \in V_-, v_i^t + \tau(v_i^h,v_j^h) \leq v_j^t\bigg\}$.
The travel-time constraint filters out edges where a relocating driver from
    supply excess node cannot reach the deficit node in time.
Each edge $e_{ij}$ is associated with utility:
    \begin{equation*}
        \mathcal{U}_{(i,j)} = \underbrace{Q_I(v_j^t, v_j^h ,v_j^h )}_\text{wait
            action at $v_j^h$} - \underbrace{cost(v_i^h,v_j^h)}_\text{relocation cost} -
    \underbrace{Q_I(v_i^t, v_i^h, v_i^h).}_\text{wait action at $v_i^h$}
    \end{equation*}
Thus, the utility of an edge measures the net value for a driver relocating along it
    during coordinated behavior.

\spara{Rebalancing operation:}
Given a rebalancing graph {\rebalancinggraph}, we wish to relocate drivers from
    supply excess zones to supply deficit zones.  
We aim to find a matching that maximizes the net reward of all relocations, 
    in order to maximize the driver earnings.  
Such a rebalancing operation can be achieved by solving a
    \emph{Minimum Cost Flow} problem expressed in form of the linear program below.
    \begin{eqnarray*}
        \text{maximize } &\sum_{e_{ij} \in E} f_{ij} \times \mathcal{U}_{(i,j)} \nonumber \\
        \text{s.t., }& \nonumber\\
        \forall e_{ij} \in E, &f_{ij} \geq 0 \nonumber\\
        \forall v_i \in V_+, &\sum_{v_j \in V_-} f_{ij} \leq \delta(v_i^t, v_i^h)
        \nonumber\\
        \forall v_j \in V_-, &\sum_{v_i \in V_+} f_{ij} \leq |\delta(v_j^t, v_j^h)|
    \end{eqnarray*}
\noindent Here, we calculate the number of
    excess drivers who should relocate from an excess node to a deficit node and store
    it in form of a flow vector $f \in \relocatingdrivers^{|E|}$ indexed along the
    edges set such that $f_{ij}$ denote the flow from $v_i$ to $v_j$. 

If the platform aims to maximize demand fulfillment metric, we can formulate it
    as a \emph{Maximum Flow} problem. 
This can be achieved by setting the utility associated with each edge 
    $\mathcal{U}_{(i,j)}=1$.

As the constraint matrices -- in both problems -- are unimodal, the solutions 
    of the linear programs are integral flow vectors and are thus optimal. 
Note that the size of the constraint matrix increases with a decrease in 
    the {\imbalancethreshold} parameter. 
However, we can greatly reduce the sizes of corresponding linear programs and 
    hence the computation time by solving a set of smaller linear programs; 
    one for each connected component of the rebalancing graph.

\subsubsection{Coordinated learning}
Based upon the computed imbalance matrix ({\imbalancematrix}) and the solution
    to the rebalancing operation above, we are now in a position to
    update the coordinated learning matrix ($Q_C$) and the degrees of coordination
    ({\coordination}) as described below.
It should be noted that while the choice of coordinated action from the matrix 
    $Q_C$ is
    influenced by the reward-maximizing rebalancing described above, the degree
    of coordination {\coordination} is merely influenced by the supply-demand
    imbalances induced as a result of the policy learnt so far.

\spara{Updating $Q_C$ for rebalancing operation:}
We capture the rebalancing operation in form of a rebalance matrix 
    $\rebalancematrix \in \mathbb{R}^{|\timesteps|
    \times m \times m}$ where each entry $\zeta(t, h, h')$ denotes a probability of
    a rebalancing relocation from zone $h$ to zone $h'$ being required at time $t$.
For every edge $e_{ij} \in E$, we update {\rebalancematrix} as follows,
    \begin{eqnarray*}
        \zeta(v_i^t, v_i^h, v_i^h) &=& 
            \frac{\delta(v_i^t, v_i^h) -\sum_{v_j \in V_-} f_{ij}}{\delta{v_i^t,
            v_i^h}} \nonumber\\
        \zeta(v_i^t, v_i^h, v_j^h) &=& \frac{f_{ij}}{\delta(v_i^t, v_i^h)}.
    \end{eqnarray*}
Using the rebalance matrix, we update $Q_C$ in line 18 of
    Algorithm~\ref{alg:general_learning} as follows,
    \begin{equation}
    \label{eq:Q_C_update}
        Q_C(t, h, h') \leftarrow (1 - \alpha)Q_C(t, h, h') + \alpha \zeta(t,
        h, h').
    \end{equation}

\spara{Updating degree of coordination ({\coordination}):}
At the end of each training episode $(k+1)$, we use the realized imbalance
    matrix ({\imbalancematrix}) to determine the degree of coordination required 
    within each zone at
    every time step.  
We update the degree of coordination as follows:
    \begin{equation}
    \label{eq:doc_update}
        \coordination_{k+1}(t, h) = (1 - \alpha)\coordination_{k}(t, h) + \alpha\mu(t, h),
    \end{equation}
    where the rebalancing raio $\mu$ is computed as:
    \begin{equation*}
        \mu(t, h) = 
        \begin{cases}
        \frac{\delta(t,h)}{S^{\policy_k}(t, h)} &\text{if }
            \delta(t,h) > 0. \\ 
            \frac{\big|\delta(t, h)\big|}{\sum_{h'}d_t(h,h')} &\text{if }
            \delta(t,h) < 0\text{ and } \coordination_{k}(t, h) > 0. 
        \end{cases}
    \end{equation*}
While the former condition encourages driver relocations in zones with supply excess, 
    the latter condition discourages it in zones with supply deficit.
Thus, we use Equation~\eqref{eq:doc_update} to update the degree of coordination
    for each zone in line 17 of Algorithm~\ref{alg:general_learning}.

\subsection{Discussion}

We conclude this section by highlighting some of the features of our approach 
    that make it appealing to use in practice. 

First of all, over the course of training, our approach learns by trialing driver
    actions over historically observed demand data and recommends 
    strategic relocations to drivers when there is enough evidence to do so.
This is done \emph{proactively}, i.e., with the goal of preventing such an
    imbalance from actually occurring. 
One should contradict this with other reinforcement
    learning based approaches~\cite{Lin2018-vs, Li2019-bp} that try to resolve the
    imbalance issues \emph{ex-post}, 
    or dynamic pricing based approaches which assume full knowledge of future
    demand~\cite{Ma2018-hb}.

On deployment, our model relies only on trends learned from the 
    historical data.
This design decision is motivated by the empirically observed strong periodicity in demand.
It makes our model relatively parameter-free, thereby providing robustness to
    demand perturbations.
This decision is validated by the model generalizability experiment in
    Section~\ref{sec:model_generalizability}, where we show that the recommendations
    made by our algorithm are robust to the presence of perturbations in the demand.
In contrast, deep reinforcement learning based approaches~\cite{Mnih2013-sj,Tang2019-xu,Lin2018-vs, Wen2017-vp, 
    Wang2018-bv} require as inputs full knowledge of supply and demand 
    distribution during deployment. 
The stochastic
    gradient-descent algorithm used during training of networks uniformly
    samples experiences observed under previous training policies, making it
    impossible to reliably trace back and explain the actions recommended
    to the drivers.
Furthermore, this increases the sensitivity of model 
    performance to the tuning of numerous hyperparameters, making them 
    difficult to deploy in the real world.

A key characteristic of our approach is the explicit coordination achieved
    by solving a \emph{minimum cost flow} problem, allowing all our
    recommendations to be easily traced back and explained.

%% file: algorithm.tex
\begin{algorithm}
Initialization $Q_I(t, h, a) \leftarrow 0, Q_C(t, h, a) \leftarrow 0, \coordination(t, h) \leftarrow 0$\;
\For{each episode $e = 1, \cdots, E$}{
    \For{each time step $t = 1, \cdots, T$}{
        \For{each driver $i = 1, \cdots, n$}{
            Generate two random numbers $\eta_0, \eta_1 \in [0,1]$\;
            \uIf{$\eta_0 \leq \epsilon$}{
                Choose exploratory action\;
            }
            \Else{
                \uIf{$\eta_1 \leq \coordination(t, h_i)$}{
                $a$ = Independent action $a^*$ from $Q_I$\;}
                \Else{
                $a$ = Coordinated action $a^c$ from $Q_C$\;
            }
            }
            Receive reward $E(t, h_i, a)$\;
            Compute rebalance matrix $\bf Z$\;
        }
    }
    \For{each zone $h \in \hexbins$}{
        $\forall t, a$ update $Q_I(t, h, a)$ \;
        $\forall t$ update degree of coordination $\coordination(t, h)$ \;
        $\forall t, a$ update $Q_C(t, h, a)$ \;
    }
}
\caption{General learning approach}
\label{alg:general_learning}
\end{algorithm}

%% file: experiments.tex
\section{Data and Experiments}
\label{sec:data_and_experiments}

In this section, we begin by describing the
    pre-processing we did in order to use the New York City 
    Yellow taxi rides public dataset and then we
    evaluate our framework. 

\subsection{Data pre-processing}

To train our model, we need to construct the time-evolving city
    matrices - {\demandmatrix}, {\rewardmatrix}, and {\traveltimematrix} 
    described in Section \ref{sec:problem_setup}.

\spara{Hexagonal binning of New York City:}
We employ the popular methodology of hexagonal binning to discretize the city 
    into a set $\hexbins$
    of 250 non-overlapping uniform-sized hexagonal zones. 
The distance from the center of a zone to its vertices is about 1 mile.

\spara{Forming time-evolving matrices:}
We begin with the NYC Taxi dataset (2015), which contains
    street-hail records of over 200,000 taxi rides per day with information
    regarding pickup and dropoff locations and times, fare, trip distances, etc.,
    from before the significant confounding effects of ride-sharing
    platforms like Uber, Lyft, etc. 
For each ride in the dataset, we evaluate its pickup and dropoff zones
    based on location coordinates. 
Assuming that passengers do not
    hail a taxi for short distances, we ignore a small percentage of rides
    which begin and end within the same zone.

We discretize a 24-hour day into 288 time-slices of
    duration 5 minutes each, indexed by their start time. 
Thus, to populate the entries of the matrices {\demandmatrix}, {\rewardmatrix}
    and {\traveltimematrix} at time $t$, we use the rides from the dataset in the 5
    minutes time-slice beginning at time $t$. 
Due to variations in the popularity of particular pickup and dropoff
    zones at specific times of the day, the {\rewardmatrix} and
    {\traveltimematrix} matrices obtained using this
    method are sparse. 
However, to compute the best policies, our framework requires 
    the availability of complete information regarding rewards and travel times.
Hence, we estimate the missing values in these matrices using linear regression
    models including fixed-effects for the time of the day, the source and
    destination zones\footnote{To compute the travel time entry
    $\tau(i,j)$ at time $t$ on a Monday, we fit a linear regression model $\tau(i,j,
    t) = \overline{\beta_0} X_{i,j,t} + \beta_1 \alpha_i + \beta_2 \alpha_j + \beta_3 \alpha_t + \epsilon_{i, j, t}$ where
    $X_{i,j,t}$ are the time-variant independent variables, the $\alpha_i,
    \alpha_j$, and
    $\alpha_t$ are time-invariant fixed-effects variables for source, destination and
    time of the day respectively, while $\epsilon_{i,j,t} \sim \mathcal{N}(0,1)$ is
    an error term. We fit this model to data from all the rides in
    the corresponding month.}.
The performance of our model is not sensitive to the
    choice of a specific linear regression modeling technique.

\subsection{Experimental results}

\subsubsection{Settings} For all experiments, we use a multiprocess implementation of our algorithm
    on a 24-core 2.9 GHz Intel Xeon E5 processor with 512 GB memory. 
The model training time for 100 episodes of training takes less than an hour. The
model testing time is less than 5 minutes. 
Our code has been made publicly available for reproducibility
purposes~\cite{github-page}.
All our experiments use learning rate $\alpha= 0.01$ and 
    discount factor $\gamma= 0.99$.
During independent learning, the exploration factor ($\epsilon$) used 
    in $\epsilon$-greedy Q-learning decreases exponentially as the 
    training progresses.
Unless mentioned otherwise, we train 5,000 drivers over 200 episodes 
    and set the imbalance threshold ({\imbalancethreshold}) to 2. 
Experimental results presented in this paper are obtained by training models 
    over a representative day viz., first Monday of September 2015 
    with a demand of over 232,000 rides.
However, our results generalize to every day in the month.

\subsubsection{Model performance}
\begin{figure}[ht]
	\centering
    \includegraphics[width=8cm, height=4cm]{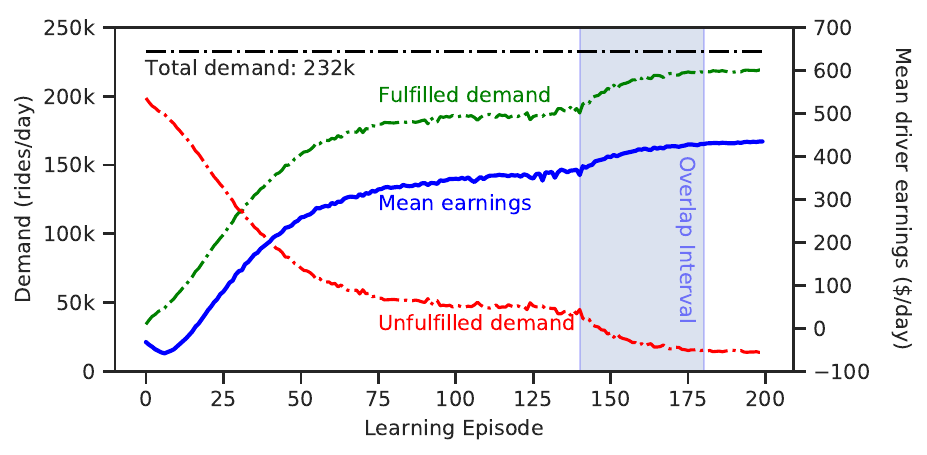}
	\caption{A representative illustration of improvement in mean driver 
    earnings during training.} 
	\label{fig:rl_training}
\end{figure}

First, we address the question: \textit{how well does our reinforcement
    learning-based algorithm learn the driver dispatch policy?}
In Figure~\ref{fig:rl_training}, we observe the improvement in mean driver 
    earnings and demand fulfillment as the training progresses. 
We split the 200 training episodes into independent learning episodes
    ($E_{IL}=160$) and coordinated learning episodes ($E_{CL}=60$). 
This can be achieved by setting the degree of coordination ($\xi$) to 1 until 
episode number $E - E_{CL}$ on line 9 of Algorithm~\ref{alg:general_learning}.
Consequently, episodes $[140, 160]$ utilize both independent and
    coordinated learning. 
In Figure~\ref{fig:rl_training}, we
    observe a significant improvement in the objective in the interval 
    denoted by a shaded region. 
As expected, coordinated learning appropriately relaxes some of the constraints imposed by 
    single-agent MDP and leads to significantly better performance.

\begin{figure}[ht]
    \centering
    \begin{tabular}{ c }
        \includegraphics[width=8cm, height=4cm]{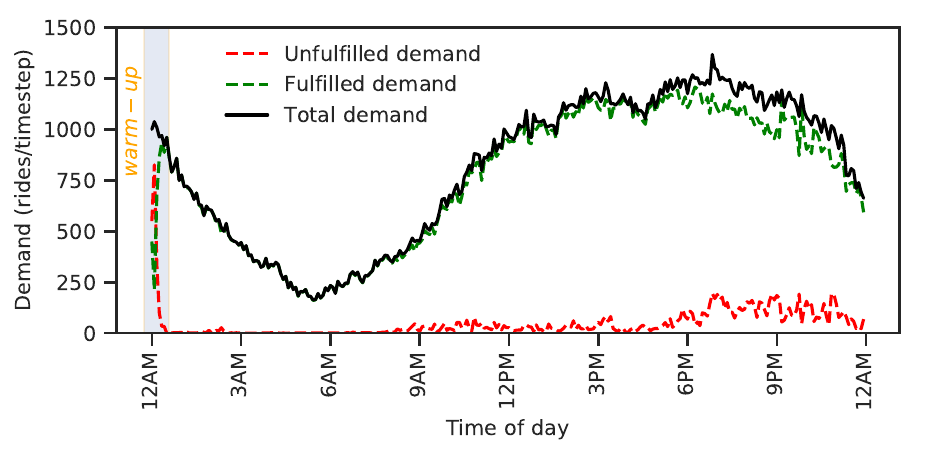} \\
        \includegraphics[scale=0.5]{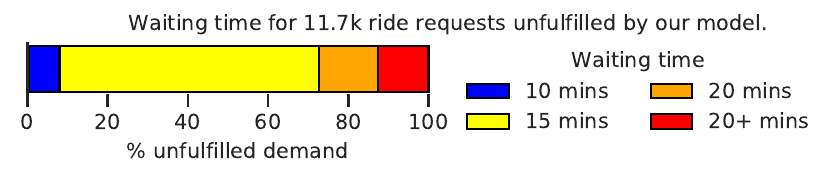}
    \end{tabular}
    \caption{Top: Demand fulfillment by a trained policy at different times on a representative
        day. Bottom: Waiting times for demand not immediately fulfilled by the model.}
    \label{fig:model_performance}
\end{figure}

In Figure~\ref{fig:model_performance}, we plot the total demand at various 
    times in the day, along with its fulfilled and unfulfilled portions
    by drivers following our policy. 
About 95\% of the total demand during the day is
    satisfied with our framework. 
We consider a ride request fulfilled if an idle driver is present in the 
    same zone at the time of the request. 
We find that 10\% of the unfulfilled demand can be fulfilled by drivers nearby 
    within 10 minutes of passenger wait while over 70\% within 15 minutes.
At the beginning of a day, for lack of better alternative, we initialize drivers 
    uniformly across the city zones. 
Hence, our model requires a ``warm-up'' time for the drivers to reposition
    themselves in order to fulfill the demands. 
This warm-up interval contributes significantly to the unfulfilled
    demand at the beginning of the day from 12AM-1AM. 
One may left-pad the training interval to alleviate this issue.

\begin{figure*}[h]
    \centering
    \includegraphics[width=15cm, height=5cm]{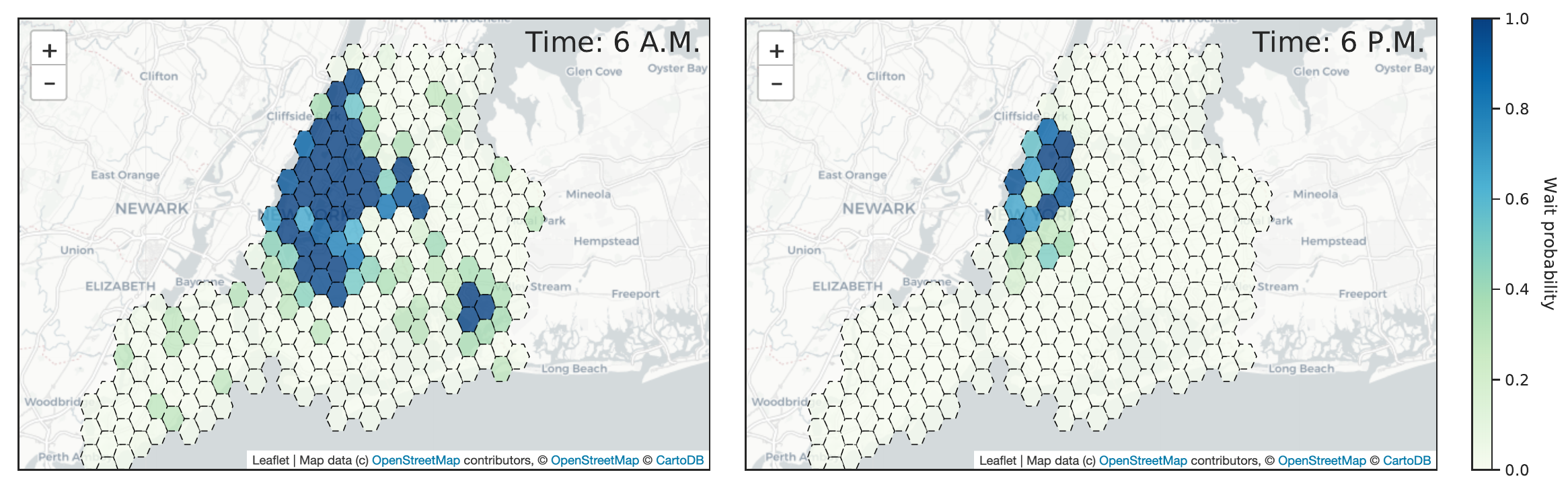}
    \caption{
        Heatmaps of probability of coordinated wait action i.e., $Q_C(t, h, h)$ during
        morning commute at 6 A.M. (left) and evening commute at 6 P.M. (right).
        }
    \label{fig:wait_probability}
\end{figure*}

The explicit coordination in our model 
    allows us to visualize the market conditions in which it is utilized.
For brevity, in Figure~\ref{fig:wait_probability}, we plot snapshots of
    coordination in form of a heatmap with probabilities of 
    coordinated wait actions i.e. $Q_C(t, h, h)$ at 6 A.M. during the early 
    morning commute and at 6 P.M. during the evening commute hours\footnote{More
    detailed visualizations depicting evolution of coordinated actions and degree of
    coordination across the city and through the time of the day are available 
    at~\cite{github-page}.}.
Without coordination, we would expect all the drivers in the city to relocate
    to Manhattan in order to satisfy the extremely high volume of demand during
    the morning commute. 
However, as observed in Figure~\ref{fig:wait_probability},
    our model recommends a certain proportion of drivers to wait in the outer
    boroughs of New York City for the early morning commute to Manhattan.
Notably, the model is able to learn demand trends in time-dependent hotspots
    such as the J.F.K. airport to the south-east of the city.
In contrast, during the evening commute to outer boroughs, the model exceedingly recommends
    that the drivers wait inside Manhattan.

\subsubsection{Impact of independent and coordinated learning}
\begin{figure}
	\centering
    \includegraphics[scale=0.5]{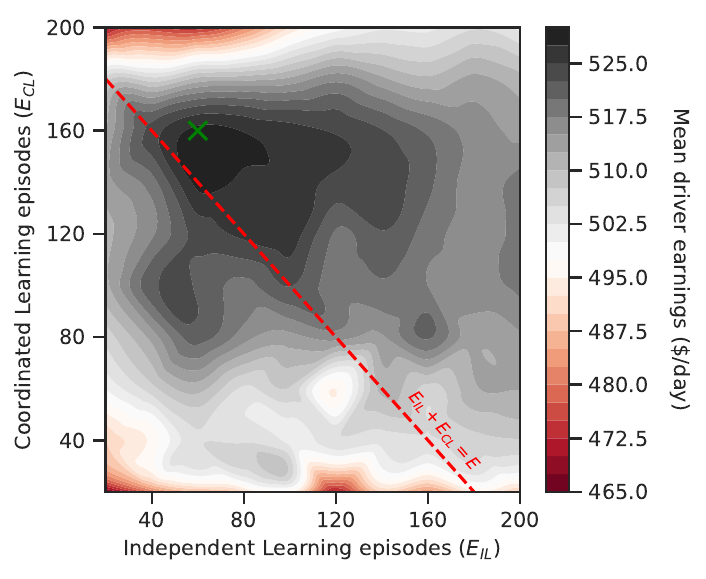}
    \caption{Performance stability over a wide range of overlaps between the
    independent and the coordinated learning.}
	\label{fig:il_cl_overlap}
\end{figure}

The overlap between the independent 
    and the coordinated learning during training is a crucial 
    aspect of our framework.
In this section, we address the question:
    \textit{how do we determine
    the appropriate number of independent learning and coordinated learning
    episodes during training?}
Given a fixed number of training episodes $E$, we assume that 
    our model trains the initial $E_{IL}$ episodes with independent learning 
    and the final $E_{CL}$ episodes with coordinated learning. 
When $E_{IL} + E_{CL} \geq E$, we have $E_{IL} + E_{CL} - E$ episodes of overlap
    between independent and coordinated learning.
In Figure~\ref{fig:il_cl_overlap}, we use 200 episodes of training, and 
    we vary the values of $E_{IL}$ and $E_{CL}$ in the range $[20,200]$ to
    achieve various overlaps\footnote{Note that there is no overlap between 
    the independent and the coordinated learning phases
in the lower triangle of Figure~\ref{fig:il_cl_overlap} when $E_{IL} + E_{CL} < E$.}.
We then plot the mean driver earnings for each learned policy.
We show that for a large interval of values of $E_{IL}$ and $E_{CL}$, 
    our framework provides stable and high performance with up to
    \$535 mean earnings per day when $E_{IL}=60$ and $E_{CL}=160$,
    denoted by a green marker in the figure.
This observation supports our claim that our framework is robust to
    imperfections in hyperparameter tuning.
Note that we have used different values of $E_{IL}$ and $E_{CL}$ in
    Figure~\ref{fig:rl_training} in order to clearly portray the incremental
    impact of coordinated learning on mean driver earnings per day.

\subsubsection{Impact of Driver Supply} 

We next answer the question: \textit{what is an appropriate number of drivers
    to fulfill the ride demand?}
To study this question, we vary the driver supply in the
    range $[1000, 6000]$, where the units are individual drivers. 
Given a fixed supply size, we plot the ratio of the number of successful driver 
    waits resulting into passenger rides to the number of unsuccessful driver 
    waits while taking into account the overall demand fulfillment.
When the number of drivers is small compared to the demand, the
    drivers should have an easier time finding a passenger. 
On the other hand, a city saturated with drivers should result in a 
    higher number of unsuccessful driver waits.
\begin{figure}
	\centering
	\includegraphics[width=8cm, height=4cm]{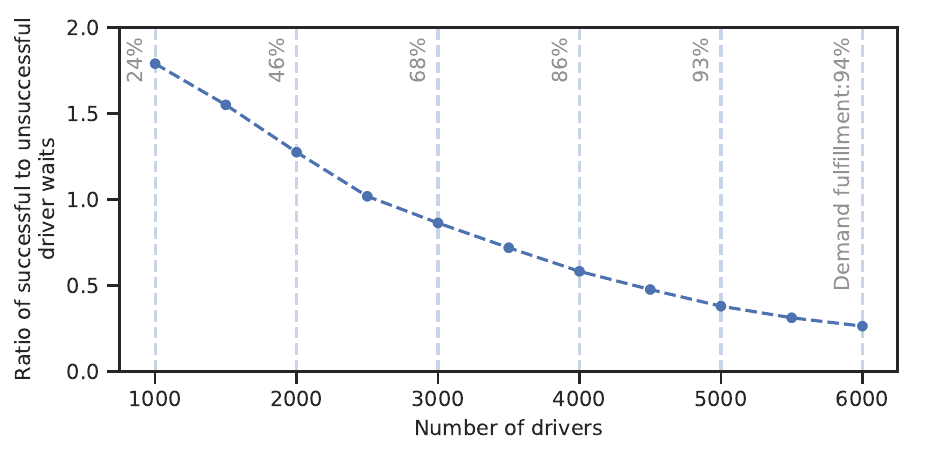}
	\caption{Impact of supply size on the ease of finding a
    passenger on a representative day.}
	\label{fig:number_of_drivers}
\end{figure}
In Figure~\ref{fig:number_of_drivers}, we observe that the framework validates
    our expectations.
The ``warm-up'' period described in Figure~\ref{fig:model_performance}
    causes underestimation of demand fulfillment while simultaneously, 
    it causes overestimation of 
    the number of unsuccessful driver waits. 
This experiment provides evidence that over 93\%\footnote{Excluding the
    warm-up interval from the analysis improves the demand fulfillment to 96\%.} demand of 
    New York City can be fulfilled by about 5,000 drivers. 
Note that in September 2015, New York City had over 
    13,500 operational taxicab medallions~\cite{wiki-001}. 
It also justifies our decision to use 5,000 drivers in most experiments 
    described in this work.

\subsubsection{Impact of platform objectives}
So far, our experiments focused on the \textsc{MaxEarnings} problem.
A natural question is: \textit{should a platform optimize driver dispatches to maximize their earnings 
    or to maximize demand fulfillment?}
Note that while maximizing the demand fulfillment might help retain customers 
    over a longer-term, it can
    be detrimental to drivers' earnings.
\begin{figure}
	\centering
	\includegraphics[width=8cm, height=4cm]{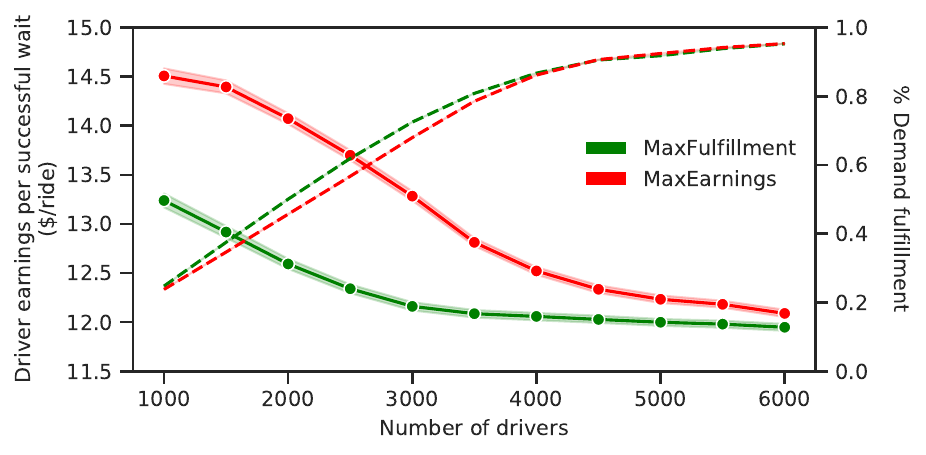}
    \caption{Differential impact of platform objectives.
    Left Y-axis: Driver earnings per passenger ride 
    with 90\% confidence interval.
    Right Y-axis: \% Demand fulfillment for both objectives on a 
    representative day.}
	\label{fig:pickups_vs_revenue_objective}
\end{figure}
To solve \textsc{MaxFulfillment} (see Section~\ref{sec:problem_setup}),
the framework rewards (resp.\ penalizes) a
successful passenger pickup (resp.\ unsuccessful wait) by +1 (resp.\ -1)
net reward.
Figure~\ref{fig:pickups_vs_revenue_objective} depicts that mean driver earnings 
    per passenger ride can be over a \$1 lower in a policy optimized for 
    maximizing demand fulfillment relative to one optimized for earnings.
The additional rides covered by the solution to \textsc{MaxFulfillment} may  
    direct drivers to sub-optimal locations
    and compromise their future earnings for the day.
As the supply increases over the minimum number of required drivers, the 
    two objectives converge while a statistically-significant difference in 
    the driver earnings per ride persists.
Note that higher rewards/penalties while solving 
    \textsc{MaxFulfillment} result in larger divergence between the 
    two objectives.

\subsubsection{Advantage of strategic behavior}
Next, we address the question: \textit{does our model provide consistently
    higher earnings for all the drivers?}
To explore this, we model the taxi driver population of the city as comprised
    of strategic drivers who follow the model recommendations and naive drivers
    who act upon heuristics learned via experience. 
We expect the mean earnings of drivers to decrease as the number of
    strategic drivers on the platform increases.
While modeling a naive driver, we assume that taxi drivers, over time, 
    learn the popular spots in the city where they are more likely to find 
    a passenger. 
If they are unable to locate passengers reasonably quickly 
    in other parts of the city, they head back to the popular spots. 
    We designate 15 zones 
    as
    popular zones based on the historical demand data.
Furthermore, we assume that an idle naive driver looking for a passenger decides to 
    head back to one of the popular zones with a fixed probability of 0.25. 
Upon choosing to relocate, the naive driver picks the target popular zone with a 
    probability inversely proportional to its distance from the current location.

\begin{figure}
	\centering
	\includegraphics[width=8cm, height=4cm]{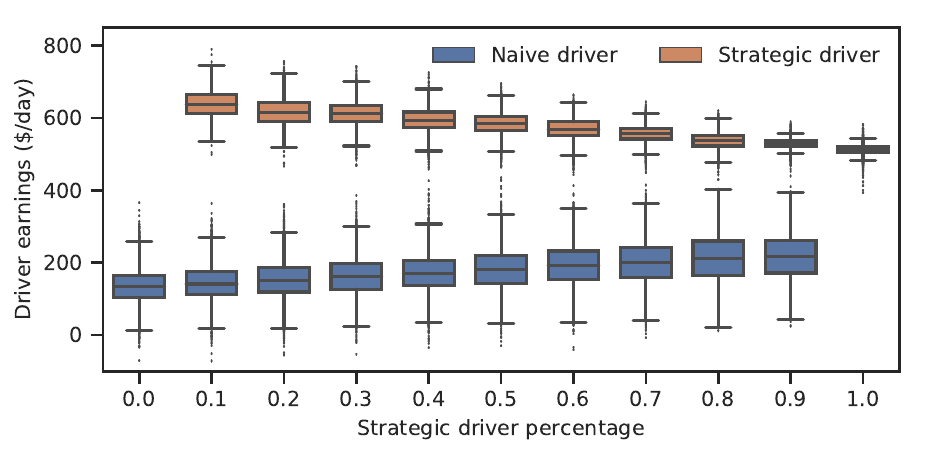}
    \caption{Earnings advantage of the strategic drivers over the naive drivers 
    on a representative day.} 
	\label{fig:strategic_vs_naive_driver}
\end{figure}

In Figure~\ref{fig:strategic_vs_naive_driver}, we plot the earnings of
    the two categories of drivers while varying the percentage of
    strategic drivers\footnote{The lower and upper edges of the boxes in
    Figure~\ref{fig:strategic_vs_naive_driver} indicate quartiles Q1 and Q3
    respectively, and the length of whisker is 1.5 times IQR.}. 
As expected, an increase in the number of strategic drivers, causes their individual 
    earnings to decline. 
Overall, the strategic drivers not only earn more than the naive
    drivers, but also the variance in their earnings is significantly lower.
Thus, our framework is \emph{envy-free} i.e., drivers at same
    location and time do not envy each other's future earnings.

\subsubsection{Model generalizability}
\label{sec:model_generalizability}
\begin{figure}[ht]
	\centering
    \includegraphics[width=8cm, height=4cm]{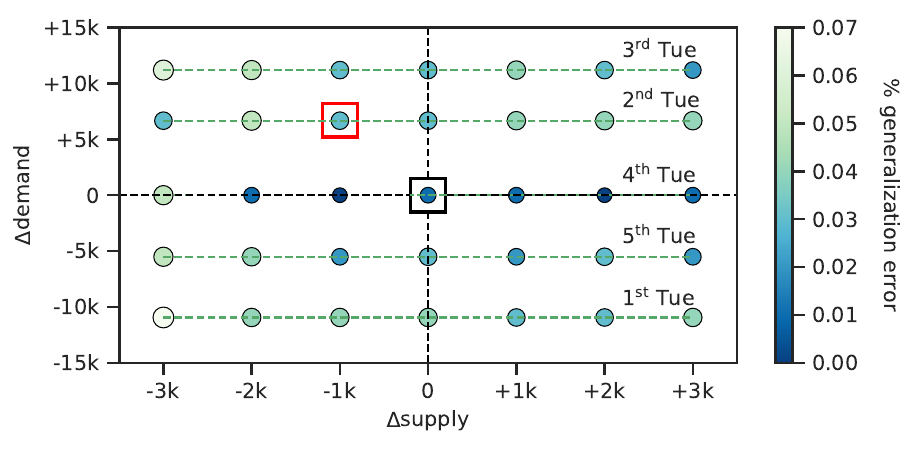} 
    \caption{Model robustness: Baseline model 
    (enclosed within a black box at origin) is 
    deployed on other Tuesdays.} 
	\label{fig:model_generalizability}
\end{figure}
In this section, we explore the question of model generalizability: 
    \textit{does our
    model perform well when deployed on days with considerably different supply-demand conditions 
compared to the day it was trained on?}
We cross-validate our model by evaluating the policy of a trained model on different days.

For illustrative purposes, we choose as baseline --
    $m_0$ -- a model
    trained to satisfy the demand of 288,000 rides observed on
    the fourth Tuesday of September using 7,000 drivers. 
We test the policy $\pi(m_0)$ recommended by our 
    baseline model by deploying it on other Tuesdays of the month. 
Note that the observed demand, as well as the number of active drivers might
    vary on other Tuesdays compared to
    our baseline model. 
To capture this potential for change in supply, we vary the number of
    simulated drivers during testing in the range $[4000, 10000]$.
In Figure~\ref{fig:model_generalizability}, enclosed within a red square box is
    an illustration of the generalization error associated with deploying our
    baseline model's recommended policy on the second Tuesday of the month with
    just 6,000 drivers.
Importantly, the policy $\pi(m_0)$ now attempts to fulfill an increased 
    demand of about 7,000 extra rides ($\Delta demand$) using 1,000 fewer drivers 
    ($\Delta supply$) than it was trained for.
To evaluate its performance in this task, we compare it with a model
    $m_*$ which was explicitly trained to fulfill the demand of
    the second Tuesday with exactly 6000 drivers.
Thus, we compute the baseline policy's generalization error as
    \begin{equation*}
        \% \textrm{generalization error} =
        \frac{\mathcal{F}^{\pi}(m_*)-\mathcal{F}^{\pi}(m_0)}
        {\mathcal{F}^{\pi}(m_*)},
    \end{equation*}
    where $\mathcal{F}^{\pi}(m)$ denotes the demand fulfillment (\%) of the model $m$.

Figure~\ref{fig:model_generalizability} shows that our framework
    generalizes well to perturbations in both supply and demand. 
We also observe that decreasing the number of drivers excessively 
    impacts harms its generalization performance. 
As a result, we recommend deploying models trained with a reasonably higher 
    number of drivers than minimally required so that they generalize better in 
    cases of varying demand. 
For brevity, we have presented a single illustrative example here; 
    the generalizability result holds true across all the models.

\subsubsection{Comparison with baselines}
\label{sec:comparative_baselines}
\begin{figure}
	\centering
	\includegraphics[width=7cm, height=4cm]{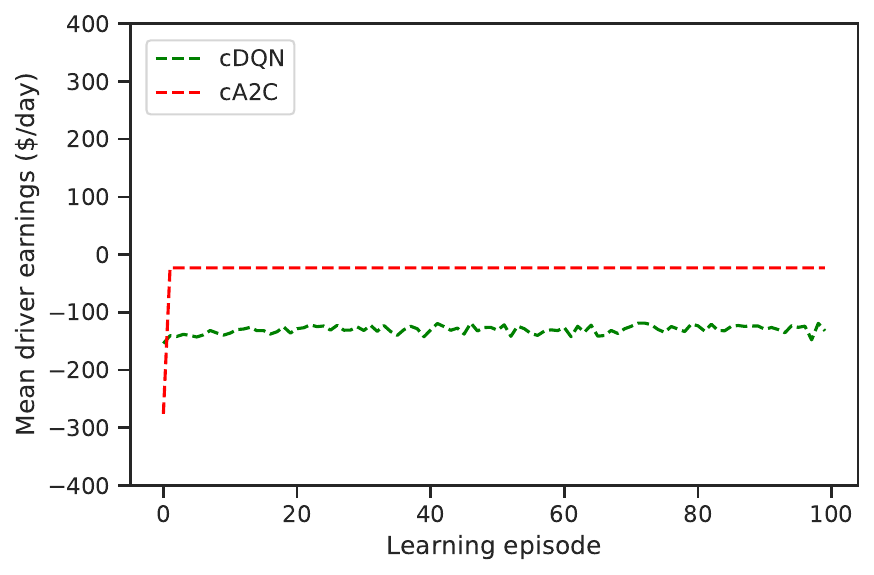}
    \caption{Performance of cDQN and cA2C deep-learning approaches from ~\cite{Lin2018-vs}.} 
	\label{fig:comparative_baselines}
\end{figure}
A major challenge in comparative studies in this domain is the
    lack of reproducibility due to proprietory datasets and
    simulators.
To the best of our knowledge, although~\cite{Lin2018-vs} uses coordinated deep
    reinforcement learning approach, it is most similar to 
    ours with respect to modeling assumptions. 
In the absence of the  Didi Chuxing's proprietary driver 
    simulator and datasets direct comparison of our works is impossible. 
We make an effort to compare our approaches by re-implementing their deep 
    reinforcement learning based algorithms (cDQN and cA2C) with minimal 
    modifications to fit 
    our setting which computes future driver distributions based on simulating
    passenger pickups and dropoffs, instead of predicting them using proprietary
    models.
    
Moreover, in ~\cite{Lin2018-vs}, the authors do not train the neural network
    from its randomly initialized state.
Instead, they initialize the network based on \textit{pre-trained value networks
    based on historical means} from the aforementioned simulator.
As it is a standard practice in reinforcement learning to train the networks from
    their random initialization state and due to the unavailability of such
    pre-trained networks, our implementations attempt to learn \emph{from
    scratch}. 

Figure~\ref{fig:comparative_baselines} shows mean driver earnings per day over
    the course of model training.
As expected, even after extensive hyperparameter tuning, the baselines failed
    to learn meaningful strategies, with driver earning net negative rewards
    of -\$20 over a day.
A deeper evaluation of the baselines showed that the neural network always
    recommended a wait action in every zone throughout the day.
In the absence of a pre-trained value network, we suspect that the proposed
    algorithms are unable to explore the policy space
    effectively to learn the policies \textit{from scratch}.
We postulate that reward sharing assumption in these algorithms results in 
    superficial coordination behavior within zones causing the network to 
    fail to learn in a more realistic scenario comprising actual passenger
    pickups and dropoffs.

Our implementations of contextual DQN (cDQN) and contextual actor-critic (cA2C)
    are publicly available at \cite{github-page}. 
Our experience with training the neural network models in these baselines 
    provide a renewed appreciation of the complexity of required
    hyperparameter tuning in order to achieve maximum performance, in comparison 
    with a simple yet robust approach proposed in our work.

%% file: conclusion.tex
\section{Conclusions}
\label{sec:conclusions}

In this paper, we studied the problem of maximizing earnings of 
    drivers employed by ride-sharing platforms like Uber, Lyft, etc. 
Our work confirms the idea that even in a high-dimensional and big-data 
    domain such as ride-sharing, the inherent structure of the data can
    be leveraged to develop a simple, interpretable, fair and highly efficient
    framework that aims to achieve this goal.
Extensive simulations based on New York City taxi 
    datasets showed that our framework is easy to calibrate due to its
    robustness to imperfections in hyperparameter tuning.
Our experiments provided evidence for the differential impact of the platform's
    objectives on driver earnings.
Finally, we demonstrated that our model generalizes well to fluctuations 
    in supply and demand.
We make available an OpenAI gym environment for comparative studies.

In the future, we would like to study problems related to the impact of autonomous electric vehicles 
    on ride-sharing fleets. 
In that setting, the optimal strategies to match such vehicles to ride requests 
    would have to take into account the inherent slow battery charging processes. 
This will bring in a new \textit{charge-scheduling} aspect to an already 
    multi-faceted problem. 

%% file: acknowledgement.tex
\section*{Acknowledgements}
This research was partially funded by NSF CAREER 1253393 and NSF 1813406 awards.

%% file: main.bbl
\begin{thebibliography}{10}

\bibitem{noauthor_undated-gj}
{Business Traveller}.
\newblock {Global ride sharing industry valued at more than \$61 Billion}.
\newblock
  \url{https://www.businesstraveller.com/business-travel/2019/01/04/value-of-global-ride-sharing-industry-estimated-at-more-than-61-billion/}.

\bibitem{Yan2018-wq}
Chiwei Yan, Helin Zhu, Nikita Korolko, and Dawn Woodard.
\newblock {Dynamic Pricing and Matching in Ride-Hailing Platforms}.
\newblock October 2018.

\bibitem{Besbes2019-ds}
Omar Besbes, Francisco Castro, and Ilan Lobel.
\newblock {Surge Pricing and Its Spatial Supply Response}.
\newblock May 2019.

\bibitem{Banerjee2015-jx}
Siddhartha Banerjee, Carlos Riquelme, and Ramesh Johari.
\newblock {Pricing in Ride-Share Platforms: A Queueing-Theoretic Approach}.
\newblock February 2015.

\bibitem{Castillo2018-he}
Juan~Camilo Castillo, Daniel~T Knoepfle, and E~Glen Weyl.
\newblock {Surge Pricing Solves the Wild Goose Chase}.
\newblock March 2018.

\bibitem{Garg2019-vs}
Nikhil Garg and Hamid Nazerzadeh.
\newblock {Driver Surge Pricing}.
\newblock May 2019.

\bibitem{Lee2004-hh}
Der-Horng Lee, Hao Wang, Ruey~Long Cheu, and Siew~Hoon Teo.
\newblock {Taxi Dispatch System Based on Current Demands and Real-Time Traffic
  Conditions}.
\newblock {\em Transp. Res. Rec.}, 1882(1):193--200, January 2004.

\bibitem{Zhang2016-vz}
Rick Zhang and Marco Pavone.
\newblock {Control of robotic mobility-on-demand systems: A
  queueing-theoretical perspective}.
\newblock {\em Int. J. Rob. Res.}, 35(1-3):186--203, January 2016.

\bibitem{Seow2010-qg}
K~T Seow, N~H Dang, and D~Lee.
\newblock {A Collaborative Multiagent Taxi-Dispatch System}.
\newblock {\em IEEE Trans. Autom. Sci. Eng.}, 7(3):607--616, July 2010.

\bibitem{Xu2018-xb}
Zhe Xu, Zhixin Li, Qingwen Guan, Dingshui Zhang, Qiang Li, Junxiao Nan,
  Chunyang Liu, Wei Bian, and Jieping Ye.
\newblock {Large-Scale Order Dispatch in On-Demand Ride-Hailing Platforms: A
  Learning and Planning Approach}.
\newblock In {\em {Proceedings of the 24th ACM SIGKDD International Conference
  on Knowledge Discovery \& Data Mining}}, KDD '18, pages 905--913, New York,
  NY, USA, 2018. ACM.

\bibitem{Zhang2017-id}
Lingyu Zhang, Tao Hu, Yue Min, Guobin Wu, Junying Zhang, Pengcheng Feng,
  Pinghua Gong, and Jieping Ye.
\newblock {A Taxi Order Dispatch Model Based On Combinatorial Optimization}.
\newblock In {\em {Proceedings of the 23rd ACM SIGKDD International Conference
  on Knowledge Discovery and Data Mining}}, KDD '17, pages 2151--2159, New
  York, NY, USA, 2017. ACM.

\bibitem{Mnih2013-sj}
Volodymyr Mnih, Koray Kavukcuoglu, David Silver, Alex Graves, Ioannis
  Antonoglou, Daan Wierstra, and Martin Riedmiller.
\newblock {Playing Atari with Deep Reinforcement Learning}.
\newblock December 2013.

\bibitem{Tang2019-xu}
Xiaocheng Tang, Zhiwei~(tony) Qin, Fan Zhang, Zhaodong Wang, Zhe Xu, Yintai Ma,
  Hongtu Zhu, and Jieping Ye.
\newblock {A Deep Value-network Based Approach for Multi-Driver Order
  Dispatching}.
\newblock In {\em {Proceedings of the 25th ACM SIGKDD International Conference
  on Knowledge Discovery \& Data Mining}}, KDD '19, pages 1780--1790, New York,
  NY, USA, 2019. ACM.

\bibitem{Lin2018-vs}
Kaixiang Lin, Renyu Zhao, Zhe Xu, and Jiayu Zhou.
\newblock {Efficient Large-Scale Fleet Management via Multi-Agent Deep
  Reinforcement Learning}.
\newblock In {\em {Proceedings of the 24th ACM SIGKDD International Conference
  on Knowledge Discovery \& Data Mining}}, KDD '18, pages 1774--1783, New York,
  NY, USA, 2018. ACM.

\bibitem{Wen2017-vp}
Jian Wen, Jinhua Zhao, and Patrick Jaillet.
\newblock {Rebalancing shared mobility-on-demand systems: A reinforcement
  learning approach}, 2017.

\bibitem{Wang2018-bv}
Z~Wang, Z~Qin, X~Tang, J~Ye, and H~Zhu.
\newblock {Deep Reinforcement Learning with Knowledge Transfer for Online Rides
  Order Dispatching}.
\newblock In {\em {2018 IEEE International Conference on Data Mining (ICDM)}},
  pages 617--626, November 2018.

\bibitem{Vincent2019-eg}
James Vincent.
\newblock {AI systems should be accountable, explainable, and unbiased, says
  EU}.
\newblock {\em TheVerge}, April 2019.

\bibitem{open-ai-gym}
Greg Brockman, Vicki Cheung, Ludwig Pettersson, Jonas Schneider, John Schulman,
  Jie Tang, and Wojciech Zaremba.
\newblock {OpenAI Gym}, 2016.

\bibitem{github-page}
{GitHub Repository}.
\newblock Learn to earn.
\newblock
  \url{https://transparent-framework.github.io/optimize-ride-sharing-earnings/},
  2020.

\bibitem{Li2011-qm}
B~Li, D~Zhang, L~Sun, C~Chen, S~Li, G~Qi, and Q~Yang.
\newblock {Hunting or waiting? Discovering passenger-finding strategies from a
  large-scale real-world taxi dataset}.
\newblock In {\em {2011 IEEE PERCOM Workshops}}, pages 63--68, March 2011.

\bibitem{Yuan2011-pd}
Jing Yuan, Yu~Zheng, Liuhang Zhang, Xing Xie, and Guangzhong Sun.
\newblock {Where to Find My Next Passenger}.
\newblock In {\em {Proceedings of the 13th International Conference on
  Ubiquitous Computing}}, UbiComp '11, pages 109--118, New York, NY, USA, 2011.
  ACM.

\bibitem{Yuan2013-gj}
N~J Yuan, Y~Zheng, L~Zhang, and X~Xie.
\newblock {T-Finder: A Recommender System for Finding Passengers and Vacant
  Taxis}.
\newblock {\em IEEE Trans. Knowl. Data Eng.}, 25(10):2390--2403, October 2013.

\bibitem{Chaudhari2018-cv}
Harshal~A Chaudhari, John~W Byers, and Evimaria Terzi.
\newblock {Putting Data in the Driver's Seat: Optimizing Earnings for On-Demand
  Ride-Hailing}.
\newblock In {\em {Proceedings of the Eleventh ACM International Conference on
  Web Search and Data Mining}}, WSDM '18, pages 90--98, New York, NY, USA,
  2018. ACM.

\bibitem{Bimpikis2016-yf}
Kostas Bimpikis, Ozan Candogan, and Daniela Saban.
\newblock {Spatial Pricing in Ride-Sharing Networks}.
\newblock November 2016.

\bibitem{Ma2018-hb}
Hongyao Ma, Fei Fang, and David~C Parkes.
\newblock {Spatio-Temporal Pricing for Ridesharing Platforms}.
\newblock January 2018.

\bibitem{Tom_Suhr-ps}
Tom S{\"u}hr, Asia J.~Biega Meike, Zehlike, Krishna~P. Gummadi, and Abhijnan
  Chakraborty.
\newblock {Two-Sided Fairness for Repeated Matchings in Two-Sided Markets: A
  Case Study of a Ride-Hailing Platform}.
\newblock In {\em {Proceedings of the 25th ACM SIGKDD International Conference
  on Knowledge Discovery \& Data Mining}}, pages 3082--3092, New York, NY, USA,
  July 2019. ACM.

\bibitem{Chen2019-li}
H~Chen, Y~Jiao, Z~Qin, X~Tang, H~Li, B~An, H~Zhu, and J~Ye.
\newblock {InBEDE: Integrating Contextual Bandit with TD Learning for Joint
  Pricing and Dispatch of Ride-Hailing Platforms}.
\newblock In {\em {2019 IEEE International Conference on Data Mining (ICDM)}},
  pages 61--70, November 2019.

\bibitem{Chen2019-ko}
Haoyang Chen, Wei Wang, K{\aa}re Kjelstr{\o}m, and Emily Reinhold.
\newblock {Gaining Insights in a Simulated Marketplace with Machine Learning at
  Uber}.
\newblock \url{https://eng.uber.com/simulated-marketplace/}, June 2019.

\bibitem{Sutton1998-wd}
Richard~S Sutton and Andrew~G Barto.
\newblock {\em {Introduction to Reinforcement Learning}}.
\newblock MIT Press, Cambridge, MA, USA, 1st edition, 1998.

\bibitem{Li2019-bp}
Minne Li, Zhiwei Qin, Yan Jiao, Yaodong Yang, Jun Wang, Chenxi Wang, Guobin Wu,
  and Jieping Ye.
\newblock {Efficient Ridesharing Order Dispatching with Mean Field Multi-Agent
  Reinforcement Learning}.
\newblock In {\em {The World Wide Web Conference}}, WWW '19, pages 983--994,
  New York, NY, USA, 2019. ACM.

\bibitem{wiki-001}
{Wikipedia contributors}.
\newblock Taxicabs of new york city.
\newblock
  \url{https://en.wikipedia.org/w/index.php?title=Taxicabs\_of\_New\_York\_City},
  2019.

\end{thebibliography}
